\documentclass[manuscript]{acmart}

\usepackage{microtype}
\usepackage{balance}
\usepackage{multirow}
\usepackage{booktabs}
\usepackage{graphicx}

\AtBeginDocument{%
  \providecommand\BibTeX{{%
    \normalfont B\kern-0.5em{\scshape i\kern-0.25em b}\kern-0.8em\TeX}}}

\setcopyright{acmcopyright}

\copyrightyear{2023}
\acmYear{2023}
\setcopyright{acmlicensed}\acmConference[IDC '23]{Interaction Design and Children}{June 19--23, 2023}{Chicago, IL, USA}
\acmBooktitle{Interaction Design and Children (IDC '23), June 19--23, 2023, Chicago, IL, USA}
\acmPrice{15.00}
\acmDOI{10.1145/3585088.3589358}
\acmISBN{979-8-4007-0131-3/23/06}

\begin{document}

\title{Designing Parent-child-robot Interactions to Facilitate In-Home Parental Math Talk with Young Children}

\author{Hui-Ru Ho}
\email{hho24@wisc.edu}
\affiliation{%
  \institution{Department of Educational Psychology, University of Wisconsin--Madison}
  \city{Madison}
  \state{WI}
  \country{USA}
  }

\author{Nathan White}
\email{ntwhite@wisc.edu}
\affiliation{%
  \institution{Department of Computer Sciences, University of Wisconsin--Madison}
  \city{Madison}
  \state{WI}
  \country{USA}
  }

\author{Edward Hubbard}
\email{emhubbard@wisc.edu}
\affiliation{%
  \institution{Department of Educational Psychology, University of Wisconsin--Madison}
  \city{Madison}
  \state{WI}
  \country{USA}
  }

\author{Bilge Mutlu}
\email{bilge@cs.wisc.edu}
\affiliation{%
  \institution{Department of Computer Sciences, University of Wisconsin--Madison}
  \city{Madison}
  \state{WI}
  \country{USA}
  }
  
\acmSubmissionID{8765}


\begin{abstract}
Parent-child interaction is critical for child development, yet parents may need guidance in some aspects of their engagement with their children. Current research on educational math robots focuses on child-robot interactions but falls short of including the parents and integrating the critical role they play in children's learning. We explore how educational robots can be designed to facilitate parent-child conversations, focusing on \textit{math talk}, a predictor of later math ability in children. We prototyped capabilities for a social robot to support math talk via \textit{reading} and \textit{play} activities and conducted an exploratory Wizard-of-Oz in-home study for parent-child interactions facilitated by a robot. Our findings yield insights into how parents were inspired by the robot's prompts, their desired interaction styles and methods for the robot, and how they wanted to include the robot in the activities, leading to guidelines for the design of parent-child-robot interaction in educational contexts.
\end{abstract}



\maketitle
\begin{figure*}[h!]
  \includegraphics[height=2in]{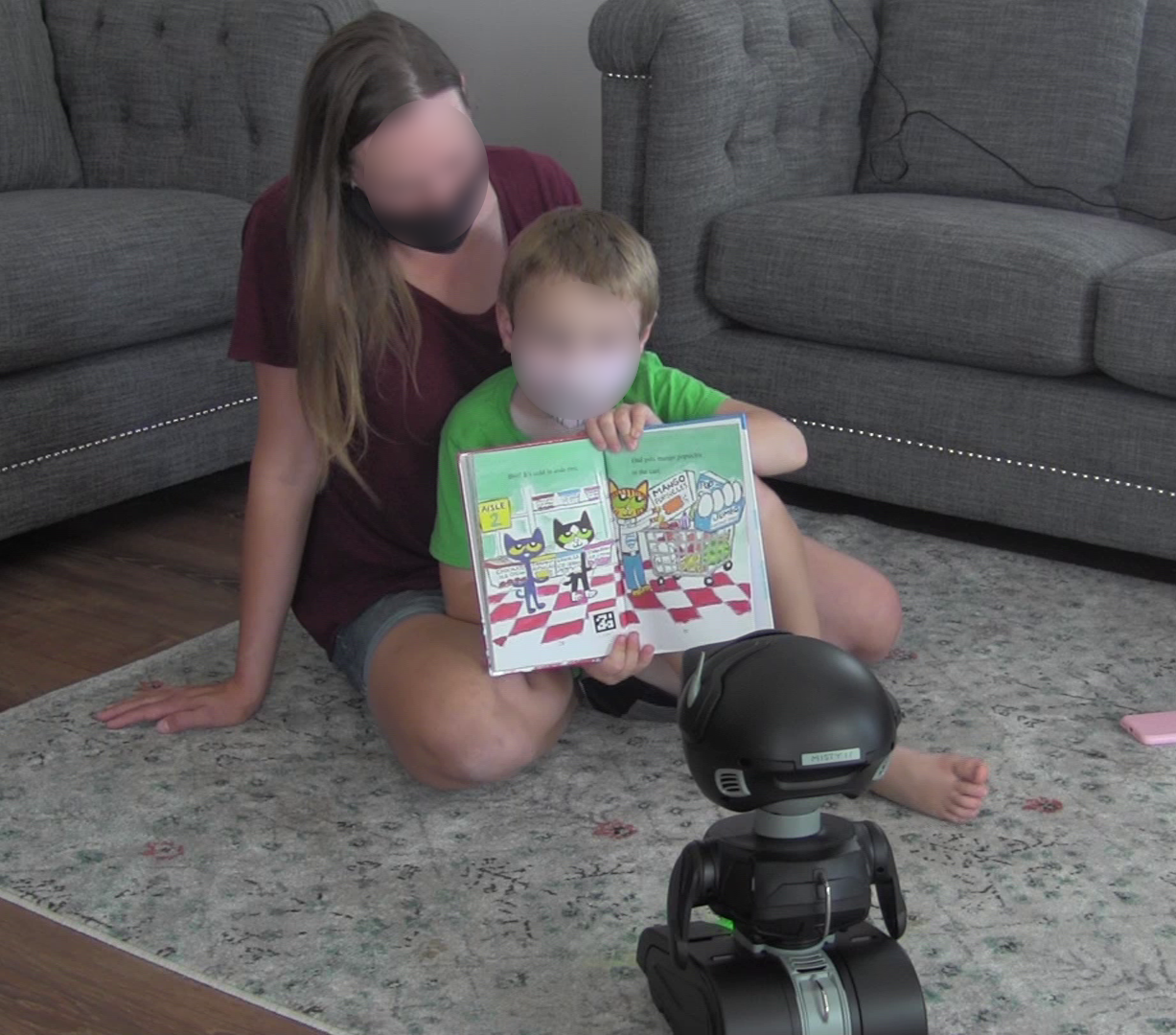}
  \includegraphics[height=2in]{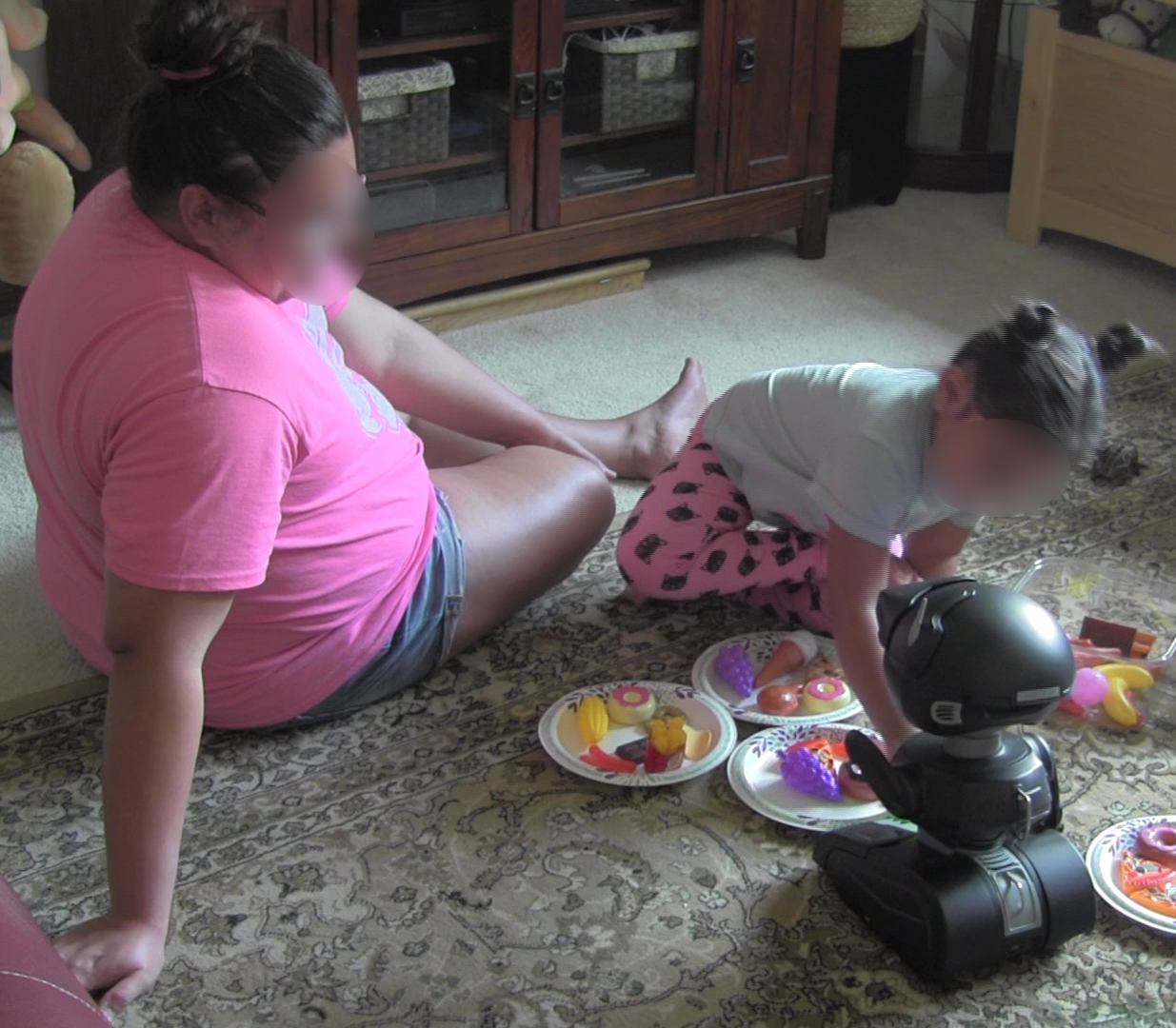}
  \caption{In this paper, we explore how families interact with a social robot that aims to prompt parents to engage in math-orientated discussions. Families read and engage in free-play with the robot in their own homes. During these activities, the robot provides math-oriented prompts related to the activity in order to promote family discussions. We explore parents' perceptions of the robot and investigate how it shapes parent-child discussions surrounding math concepts.}
  \label{fig:individual}
\end{figure*}

\section{Introduction}
Prior work has shown that mathematical competence is as important as literacy skills in predicting later educational and occupational attainment \cite{parsons2005does}. Before formal preschool education, parents play a critical role in informal learning activities, as a constructive parent-child interaction encourages young children's affective and cognitive development, including math skills \cite{levine2012early, vandermaas2012playing, susperreguy2016maternal, hart2016understanding, galindo2015decreasing}. 
Such parent-child interaction includes a concept from educational psychology called \textit{math talk}, in which parents or caretakers naturally integrate mathematical language, operations, or concepts into daily conversations with children. For example, during meal time, parents may tell the child to ``grab two apples,'' ask the child to give them ``more cookies,'' or simply prompt the child by asking ``how many hot dogs do you see?'' Parental math talk allows children to apply their math comprehension for more advanced thinking in daily-life scenarios and can predict the later math achievement of young children \cite{susperreguy2016maternal, levine2010counts, gunderson2011some, eason2020parent, eason2021facilitating}. However, due to varied demographics and socioeconomic status (SES), not all children have access to high-quality math talk with their parents \cite{dearing2022socioeconomic}.

Parents may need guidance to incorporate math talk into interactions with their child \cite{hanner2019promoting}, and social robots have the potential to help them do that. Social robots have been increasingly adopted in education, due to a variety of factors such as limited parenting resources and a need for personalized support in learning for each individual \cite{belpaeme2018social}. By facilitating physically and socially situated learning, social robots can more effectively support learning than existing technologies such as tablets and smartphones do \cite{westlund2015comparison}. 
Prior research has explored the use of social robots in supportive educational roles, acting as tutors, aides, or peers, as a way to expand educational opportunities and resources for children \cite{belpaeme2018social, cagiltay2020investigating}. 
Past research has also explored the use of social robots as educational technologies in informal learning settings, such as homes, as reading companions \cite{michaelis2018reading, michaelis2019supporting} and homework aides \cite{berrezueta2020smart}. This research has highlighted the importance of careful design considering the social dynamics of these settings, such as the involvement of the parents in the educational activity \cite{relkin2020parents, chen2022designing}.
 These studies primarily focus on child-robot dyadic interaction, but additional research should examine how social robots might also empower parent engagement in educational activities with children. Optimizing social robotic intervention to support parenting in children's math learning could have profound implications in promoting educational experiences and outcomes.

This paper investigates the role of the robot within the context of parent-child-robot interactions surrounding math talk in order to develop design implications for how social robots can support parents in their young children's learning experience. 
In the study, parent-child dyads, involving a child aged four years old, were asked to engage in reading and play activities with a social robot designed to promote math talk. We present our findings from a qualitative analysis of the recordings of the activity sessions and interviews with parents after the sessions regarding their experiences in the interaction. This work builds upon the academic literature in educational psychology and child-robot interactions to answer the following questions: 
\begin{enumerate}
     \item \textbf{RQ1:} How do parents perceive robot-guided \textit{math talk} reading and play activities?
    \item \textbf{RQ2:} How can we design a robot for parent-child-robot interactions to facilitate educational conversations?
\end{enumerate}

Our exploratory study contributes to existing research in two main areas: (1) A better understanding of how the involvement of a social robot affects parents and parent-child relationships in conversations during informal learning activities at home; (2) A set of design guidelines for an educational robot in the context of parent-child-robot interactions.

\section{Related Works}

\subsection{Math Talk}
In early home math environments, parental input plays a critical role in children's mathematical development  \cite{blevins2016early, duncan2007school, hart2016understanding}. We know that exposing children to math language at early ages positively affects their general mathematical skills \cite{purpura2017causal}, and their level of math talk predicts their understanding of the cardinal meaning of number words \cite{levine2010counts}. Thus, parents can support their child's mathematical learning at home via engaging in different kinds of activities (e.g., number board games) and incorporating math concepts in their conversations with their children. This incorporation of math concepts in at-home activities is known as \textit{math talk}. Prior studies have indicated that the frequency of children's exposure to math activities with parental participation in daily life is predictive of children's performances on a standardized test of early mathematical ability \cite{blevins1996number}. Moreover, playful activities serve as an enjoyable opportunity for parent and child to have an increased level of math talk \cite{eason2020parent}.  

Prior work has shown that parents can differ in the quantity and quality of math input they give to children at home, leading to variability in children's early math performance \cite{elliott2017understanding, thippana2020parents, eason2021facilitating}. Parents have diverse opinions regarding how important math is for their young children, and also have different subjective judgments of how proficient they themselves are at math or how much they like math. Parents' numerical approximation abilities and subjective math ability are related to their math talk \cite{elliott2017understanding}, and their attitudes also affect their ability to provide effective math talk \cite{dowker2021home}.

In addition to parental qualities, the difficulty of math talk content can influence children's learning \cite{engel2013teaching, engel2016mathematics, gunderson2011some}. One study examined the math content covered by kindergarten teachers and found that spending time on advanced math content was positively associated with learning while spending time on basic content had a negative association, suggesting that most kindergarten children will benefit from exposure to advanced content \cite{engel2016mathematics}. Another study indicated that delivering math talk involving larger number sets was more predictive of cardinal number knowledge compared to smaller sets \cite{gunderson2011some}. However, not all parents recognize the importance of discussing advanced content, thus a robot could assist them by directing their attention to environmental or contextual features associated with advanced math concepts. 

\subsection{Social Robots in Education}

Prior work has shown that social robots can positively assist in school-based learning of language \cite{kanda2004interactive, kennedy2016social, gordon2016affective, park2011teaching}, science \cite{davison2020working, shiomi2015can}, and math concepts \cite{lopez2018robotic, brown2013engaging, ramachandran2019personalized}. Similarly, there is strong evidence that in-home and informal learning companion robots can benefit children's reading \cite{michaelis2018reading, michaelis2019supporting}, comprehension \cite{yueh2020reading}, and math skills \cite{clabaugh2019long}. Social robots can support children's leaning with several capabilities such as verbal communications \cite{brown2014positive}, adaptive assistance \cite{ramachandran2019personalized}, and social interactions \cite{westlund2015comparison, lee2022unboxing}. \citet{brown2014positive} compared learning with and without a robot that provided verbal comments and found that verbal cues in robot-based learning increased student engagement and decreased boredom in math learning. \citet{ramachandran2019personalized} demonstrated that a math tutor robot could adapt its assistance to better suit each individual's needs, resulting in improved learning gains for fourth graders. Similarly, social robots have been shown to increase academic performance, user motivation, and interest in the topic \cite{lopez2018robotic, brown2013engaging} as well as offer social benefits that better engage children compared to other media. \citet{westlund2015comparison} compared children's attitudes toward three different interlocutors (human, robot, and tablet) in language learning. Their results suggest that children showed a clear preference for the robot over the tablet, perceiving it as ``someone'' instead of ``something,'' which indicates that children attended to the social cues during learning. These results suggested that children can benefit from unique features social robots afford in both formal and informal education.

 Educational robots can interact with children in various ways, from one individual to a group, involving different numbers and types of people. A good deal of studies on educational robots focus on \textit{single child-robot} interactions in the home learning environment. For example, \citet{michaelis2018reading} suggested that a social robot could maintain regular reading activity of a child (aged 10-12) at home while transforming the experience into a social one, and \citet{brown2013engaging} demonstrated that social interaction with robots help promote a child's tablet-based algebra test performance. These studies point to the potential support that social robots could offer for children in at-home education surrounding dyadic child-robot interaction design. Some other studies investigated how \textit{multiple children interact with a single robot}. \citet{strohkorb2015classification} examined social dominance in a group of children involving a social robot and identified a novel model for classifying social dominance level using non-behavioral features. On the other hand, \citet{leite2015emotional} studied how \textit{multiple robots interact with a single child} within a storytelling scenario and discovered that involving multiple robots could benefit children's social development. Other studies, yet relatively rare, involve \textit{parent-child dyads} in the interaction with a robot, leading to a triadic social dynamic. \citet{ho2021robomath} explored the design of a robot in supporting young children's math learning at home through a board game, with the presence and support of a parent. This study mainly analyzed the pros and cons of different cues from the robot and did not focus on the design for parent-child-robot interaction. Another study specifically explored how parents can support young children's learning experience with a robot in the field of programming education \cite{relkin2020parents}, but focused on highlighting parents' strategies instead of uncovering the design space for a robot in triadic interaction. 

 Recent works have begun investigating parent-child-robot interactions, but few of them focus on how a robot may influence in-home math talk between parents and preschool children, as well as the factors that must be considered when introducing a social robot designed to engage in math talk into a home environment. Prior work has shown that the backgrounds and engagement of the parents can affect the experiences children have with robots \cite{mchugh2021unusual} and that robots can improve general parent-child dialogue \cite{chan2017wakey}. In addition, \citet{chen2022designing} have explored parent-child-robot interaction design through multiple longitudinal co-reading sessions, creating design guidelines for robot-assisted parent-child interactions as well as insights into long-term experiential knowledge for technology design. However, the questions of how both parents and children perceive the role of an in-home math-oriented robot, how the robot should engage and interact when involved in activities with multiple individuals, or what primary mode of interacting it should utilize in each activity are not yet fully explored.
 


\section{Method}

To address the knowledge gap in the literature of education and human-robot interaction (HRI), we conducted an exploratory in-home field study involving parent-child activities with a robot to understand how social robots can facilitate math-focused parent-child discussions during shared activities and the perceptions parents have regarding the robot and its assistance. 

This exploratory study is a part of a larger research effort to establish the immediate and long-term effects of this educational intervention, which will involve comparing robot-facilitated math talk against naturally occurring math talk as well as studying sustained effects of the intervention after a period of the use of the robot. In this paper, we focus on a qualitative analysis of data from a small sample of participants involved in a between-participants study that asked some families to engage in reading and play activities and some families to do so with a robot. We present data only from participants who engaged with the robot to first establish an initial, qualitative understanding of family interactions with and perceptions of the robot. We plan a comparative analysis of our data from both groups following additional data collection in our future work.
 
\subsection{Participants}
Our analysis includes data from nine families with children between the ages of 4 and 5. This age group was selected as it is the time when children start to get familiar with basic math concepts \cite{leyva2021relations}. The families were recruited via email distributed through university employee mailing lists. Each study trial involved one parent and one child, although occasionally other parents and siblings were present nearby but did not participate in the activities. Participant demographics are listed in Table \ref{tab:demographics}. Each family received \$15 USD as compensation at the completion of the study.

\begin{table}[!t]
    \caption{Participant demographics}
    \label{tab:demographics}
    \centering
    \begin{tabular}{p{0.02\linewidth}p{0.1\linewidth}p{0.1\linewidth}p{0.1\linewidth}p{0.2\linewidth}p{0.2\linewidth}}
        \toprule
         ID & Child Age & Child Gender & Parent Gender & Child's Maternal Education & Household Income Level\\
         \hline
         R1 & 4 yrs 2 mos & Male & Female & Doctoral degree & \$200,000 or more\\
         R2 & 4 yrs 8 mos & Male & Male & Bachelor's degree & \$200,000 or more\\
         R3 & 4 yrs 6 mos & Female & Female & Doctoral degree & \$150,000-\$199,999\\
         R4 & 5 yrs & Male & Female & Some college, but less than 1 year & \$150,000-\$199,999\\
         R5 & 5 yrs & Male & Female & Some high school coursework or less & \$15,000-\$24,999\\
         R6 & 4 yrs 4 mos & Male & Female & Doctoral degree & \$200,000 or more\\
         R7 & 4 yrs 8 mos & Male & Female & Doctoral degree & \$150,000-\$199,999\\
         R8 & 4 yrs 3 mos & Male & Male & Master's degree & \$200,000 or more\\
         R9 & 4 yrs 2 mos & Female & Female & Master's degree & \$50,000-\$74,999\\
         \hline
    \end{tabular}
\end{table}

\subsection{Study Design}
To understand how parents perceive a robot designed to facilitate math talk with their child, we asked families to engage in 
two activities---a \textit{reading} activity and a \textit{structured free-play} activity---in which a social robot provided math talk related prompts. 
All study activities took place in the participant's home to observe parent-child interactions in their natural environment. The order of the two activities were counterbalanced across all participants.

\subsection{Study Setup}
The study materials included a video camera, an audio recorder, and materials for the parent-child activities (i.e., two storybooks, and a set of fake food toys), a Misty II social robot \footnote{\url{https://www.mistyrobotics.com/products/misty-ii/}} and ``warm-up'' booklet that we developed to familiarize families with how to interact with the robot. The video camera was placed where it could capture the interaction between the parent, the child, and all the study materials. The audio recorder and remaining materials were placed on the floor near the parent and child. The robot was placed in front of the parent and the child, as shown in Figure \ref{fig:individual}. The experimenters sat outside the view of the camera, taking notes and operating the robot.

\subsubsection{Interaction Design}
Misty II is a semi-humanoid robot with a 4-inch LCD display for its face where customizable facial expressions and animations can be displayed. During the reading and free-play activities, the robot provided children with verbal prompts and questions that were accompanied by facial expressions designed to increase engagement. The robot's speech was generated using Google's cloud text-to-speech engine\footnote{\url{https://cloud.google.com/text-to-speech/}}. Children interacted with the robot by displaying April Tags\footnote{\url{https://april.eecs.umich.edu/software/apriltag}} on books during the reading activity, placing plastic food items objects during the free-play activity, or pressing on one of two front corner bumper buttons along its base. The \textit{Repeat} bumper allowed families to indicate their desire for the robot to repeat its previous prompt. The \textit{Yes} bumper served a unique function for each activity. In the reading activity, it allowed families to confirm their book selection, while in the free-play activity it allowed families to control when the robot was allowed to give its prompt. 


\begin{table}[!t]
    \caption{Example robot prompts for both the reading and free-play activities. Each robot prompt targets one math concept.}
    \label{tab:example_prompts}
    \centering
    \begin{tabular}{p{0.2\linewidth}p{0.36\linewidth}p{0.36\linewidth}}
        \toprule
        \textbf{Math Talk Concept} & \textbf{Reading Activity Example} & \textbf{Free-Play Activity Example} \\
        \toprule
        Cardinal Values &  I think Rosie has walked five steps forward from home. & How many doughnuts do we have?\\
        \midrule
        Number Comparison & There are so many pears on the tree. Do you think the fox can eat more pears than Rosie does? & Do we have more bananas or more pizza?\\
        \midrule
        Addition & I think that if five more frogs were to come around to the pond, there would be seven frogs in total. & How does the addition change the amount of ice cream on your plate?\\\hline
        Subtraction & I wonder if the two frogs jumped away and the bird flew away, how many animals would be left around the pond? & How does the eating change the amount of ice cream on your plate?\\
        \midrule
        Spatial Language & Rosie is still in front of the fox, and now she even walked past the goat. & Are the fries inside or outside the package?\\
        \bottomrule
    \end{tabular}
\end{table}

\subsubsection{Comment Design}

For both the free-play and reading activities, we designed prompts for the robot that targeted several math talk concepts, examples of which can be seen in Table \ref{tab:example_prompts}. These concepts, from basic to advanced, include Spatial Language, Cardinal Values, Number Comparison, Addition, and Subtraction, all of which are associated with math skill development \cite{johnson2022spatial, mix2019spatial, hawes2022effects, sarnecka2013idea, le2007one, wynn1992children}. They were selected as they are typical math talk topics used in prior studies \cite{purpura2017causal, levine2010counts, susperreguy2016maternal, eason2020parent, son2020parental}, and represent a good mixture of varying levels of math concept difficulty for our age group (4--5 years old). Prompts regarding Cardinal Values focused on concrete quantities within the activity, for example, counting the number of items that appeared on a page or the number of food items placed on a plate. Number Comparison prompts attempted to elicit discussion regarding how two or more values compared to each other, whether any amount was more, less, or equal to another. Addition and Subtraction prompts attempted to elicit discussion on adding and subtracting two quantities. Finally, Spatial Language prompts attempted to elicit discussion on the spatial relationship between multiple objects, for example, whether one object was in front of another. Each of these concepts either took the form of the robot making a statement or asking a question with the goal of promoting discussion about the concept between the parent and child. 

\subsubsection{Study Activities}
We designed two activities---reading and free-play---that provided ample opportunity for parents to engage in math talk with their child in a naturalistic manner. We chose these two activities, as they were used as study tasks in prior research on the development of math concepts in children \cite{elliott2017understanding, purpura2017causal, vandermaas2009numeracy}; they represented common parent-child activities in which parents might engage in math talk; and they facilitated parent-child discussion over math concepts, e.g., through counting, adding, subtracting items on books or items in the toy set. The order of the activities was counterbalanced across participants. Prior to each activity, families read two brief ``warm-up'' booklets that provided information on how to interact with the Misty robot and that facilitated practice interactions. The booklets included guided activities that walked the families through the process of showing April Tags to the robot as well as using the bumpers to answer the robot's questions or to prompt the robot to repeat what it previously said.

\begin{figure*}[t]
  \includegraphics[width=\linewidth]{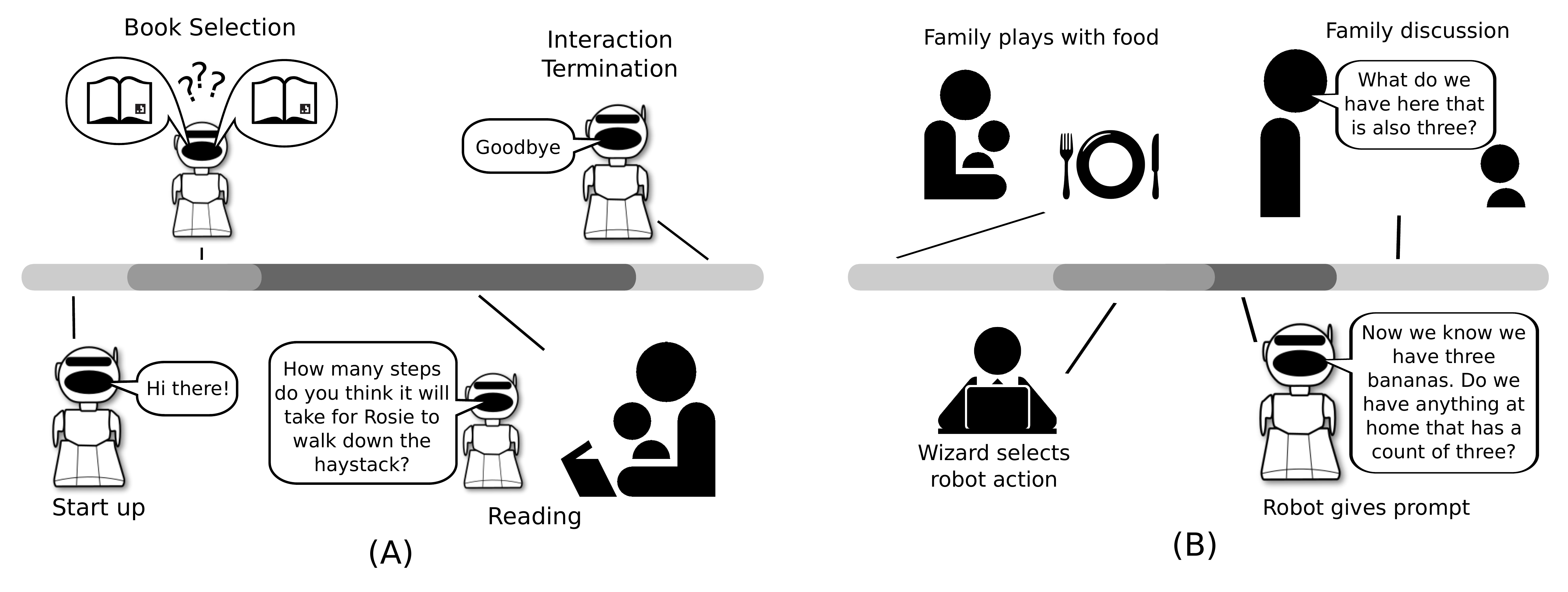}
  \vspace{-12pt}
  \caption{(A) Timeline of the reading activity. After the robot starts up, it prompts to user to select a book then give comments throughout the reading activity. (B) Timeline of the free-play activity. As families play with the toy food, the wizard selects robot actions and provides prompts for the robot to give. Both scenarios promote family discussions by providing a question or statement, upon which the parent and the child could build further discussion.}
  \label{fig:interaction_timeline}
  \vspace{-6pt}
\end{figure*}

\paragraph{Reading Activity}
In the reading activity, families were invited to choose to read either ``Pete the Cat's Trip to the Supermarket\footnote{James Dean, HarperCollins, 2019}'' or ``Rosie's Walk\footnote{Pat Hutchins, Little Simon, 2015}.'' Parents were encouraged to discuss with their children anything that came to their mind during the reading and were not asked to engage in math talk.

The books were augmented by placing April Tags on each page. These April Tags allowed the robot to autonomously recognize the given page and give an appropriate prompt that aimed to trigger math talk between the parent and child. Prior to reading, parents were informed that the interaction with the robot would be similar to the guided activity that they completed with the warm-up booklets. Parents were also told that they were free to show the April Tags to the robot whenever they wanted to during the reading activity, which allowed parents to find opportune times for the robot to speak without interrupting parent-child discussions or the reading itself. 

At the start of the interaction, the robot asked the family which book they would like to read. Following book selection, the robot asked the family to begin reading and waited for tags to be shown to it. Once the robot detected a tag, it selected an appropriate prompt for the detected page and provided the prompt to the family to engage in math talk. A timeline of the designed activity can be seen in Figure \ref{fig:interaction_timeline} (A), with examples of comments in Table \ref{tab:example_prompts}.

\paragraph{Free-play Activity}
The free-play activity was facilitated in a Wizard-of-Oz (WoZ) fashion and involved the parent and the child sharing a pretend meal using plastic food. Parents were encouraged to discuss, or expand upon, anything that came to their mind during their play session but were not explicitly asked to engage in math talk. This activity was designed to last approximately 20 minutes. Families were told that they could make and share a meal together and that they can play with the food however they wanted.

Both children and parents were guided through a pretend meal with the robot where they (1) selected food for the meal; (2) distributed the food; (3) added and distributed additional food; and (4) finally pretended to consume the food. During each phase of the activity, the robot would prompt the family to engage in different forms of math talk. Phases (1) and (2) focused on Cardinal Values and Number Comparison. Phase (3) focused on Addition, and phase (4) focused on Subtraction. Examples of prompts for each concept are provided in Table \ref{tab:example_prompts}. Note that Spatial Language concepts were not included due to the dynamic nature of the environment and the variability of object placement, making it difficult for the WoZ operator to react in real-time. To engage in the math talk, the robot raised its hand in order to indicate it wanted to speak, and then waited for families to press the \textit{Yes} bumper to allow it to provide its prompt. This behavior was designed to allow the family to determine when to engage with the robot without feeling like the robot was interrupting them. Each phase of this activity lasted approximately five minutes. At the end of each five-minute period, the robot raised its hand to ask if they were ready to continue to the next phase. If the users elected to end the phase, the robot would then instruct the users on the next phase of the activity. Otherwise, the robot gave the user more time to complete the phase before asking again.

Throughout the activity, the wizard controlled the timing of when the robot would raise its hand as well as what it would ask when the family engaged with it. The wizard also recognized the different plastic foods the family interacted with as well as the quantity of each object. A simplified timeline of the free-play activity can be seen in Figure \ref{fig:interaction_timeline} (B).

\subsection{Study Procedure}

Prior to our arrival at the participating family's home, we informed parents that they would be asked to sit on the floor during the study and asked them to clear away space. After arrival, the experimenters set up the study's recording equipment, study activities, and the robot. Following setup, the experimenters provided a high-level overview of the study and obtained informed consent: parents were asked to fill out consent forms, and the children gave verbal assent. Next, families completed the activities that were facilitated by the two warm-up booklets and the two study activities. Finally, parents participated in a semi-structured interview on their experiences, filled out a demographic questionnaire, received compensation for their participation, and signed a payment log. In total, each study ran for roughly 60 minutes, with both activities accounting for roughly 40 minutes followed by a 20-minute interview.

\subsection{Data Analysis}
For anonymization purposes, we gave each participating family an ID that represents the order they participated in the study. For example, R1 refers to the first participant.

Using the guidelines proposed by Clark and Braun \cite{clarke2014thematic}, we conducted a Thematic Analysis (TA) on the semi-structured interviews in order to analyze parents' thoughts regarding the robot and the family's interactions with it. Our thematic analysis included five stages. First, the first and the second author transcribed and familiarized themselves with the data based on the audio and video recordings. Second, the first author generated an initial codebook based on her interpretation on the data. Third, both the first author and the other experienced qualitative coder coded data from R6 with the initial codebook in order to assess and establish inter-rater alignment. Fourth, we compared the two sets of codes, discussed the differences, and adjusted the codebook as needed. Both coders coded data from R3 separately with the revised codebook and obtained an inter-rater reliability of $85\%$. Lastly, after the first author and the independent coder coded the rest of the data, the first and the second author derived, reviewed, and finally named the themes. 


\section{Results}
In this section, we present three themes from our thematic analysis of the semi-structured interviews with parents. We found that (1) parents noted that the robot broadened their awareness for how to engage in math talk, (2) parents perceived different roles for the robot and preferred for it to independently play with the child, and (3) parents want to have control over the interaction. These themes are summarized in Table \ref{tab:summary_of_analysis}.

\begin{table*}[t!]
    \caption{A summary of the results from the thematic analysis.}
    \label{tab:summary_of_analysis}
    \centering
    \begin{tabular}{p{14cm}}
        \toprule
        \textbf{Summary of Thematic Analysis} \\
        \toprule
        \textbf{Theme 1:} Parents noted that the robot broadened their awareness for how to engage in math talk
        \begin{itemize}
            \item The robot gave parents ideas for how to incorporate math talk in their parent-child activities
            \item The robot gave parents ideas for what types of math talk concepts to ask their child
        \end{itemize}\\
        \midrule
        \textbf{Theme 2:} Parents perceived different roles for the robot and preferred for it to independently play with the child
        \begin{itemize}
            \item Parents preferred that the robot would instead interact with their child independently
            \item Parents saw the robot as a teacher, playmate, and facilitator
        \end{itemize}\\
        \midrule
        \textbf{Theme 3:} Parents want to have control over the interaction
        \begin{itemize}
            \item Parents had different preferences for how the robot started each interaction
            \item Parents wanted more control over when the robot offered prompts
        \end{itemize}\\
        \bottomrule
    \end{tabular}
\end{table*}




\subsection{Theme 1: Parents noted that the robot broadened their awareness for how to engage in math talk}
\paragraph{Parents Learning from the Robot} All parents (R1, R2, R3, R4, R5, R6, R7, R8, R9) expressed positively that the robot gave them ideas for what to discuss with their child, as well as how they could incorporate those discussions into other interactions they have with their children.
R1 noted that the robot ``brought up a lot of interesting points'' and that ``it was nice to have the questions that prompted us to have a discussion.'' R1 also stated that if their child ``were to play by himself, there wouldn't have been as much counting or things.'' R5 added that the robot ``asked harder questions,'' which ``gives parents a chance to explain'' new concepts to the child. R3 felt that when interacting with their child they would always ``ask the same questions'' about the same book without the robot there to provide them with new prompts, and that the robot made them realize ``that I [R3] can be asking her [child]'' more math-related questions that R3 ``doesn't normally ask'' because she observed that her child was actually able to understand them. R6 noted the benefit to parents, saying that the robot ``provided a construct'' in a way that took some of the cognitive load off their shoulders.  R4 expressed a similar thought, saying that asking those types of questions ``would never have even cross my[R4] mind,'' and they ``hadn't really thought it could be this simple to add teaching elements'' to playful activities. 


\paragraph{Robot Word Choice} Several parents (R1, R2, R5, R8) suggested that the robot's choice of words could be more basic and straightforward for children. This desire for the occasional variation between basic and advanced terminology suggests that the robot's choice of words should depend upon the interaction dynamic. R1 provided an example, explaining that their child may understand the idea of \textit{subtraction} but ``he does not know the word,'' so R1 would clarify and exemplify it by saying ``if you have four and you take away one how many do you have'' instead. R5 also thought the robot's questions were ``a bit advanced for her child,'' and could be more ``simple and direct.'' On the other hand, R5 also expressed that the robot's use of more advanced words motivated them to engage in discussions with their child. R1 described the robot's advanced terms as a ``good teaching point,'' and stressed that if the robot would continue to use more advanced terms, she felt that she would tend to join in the activity with her child and the robot. Similarly, R5 recalled that she tried to explain the term ``even number'' to her child by saying ``here's an example you have two of these and that's an even number,'' showing that the robot was able to motivate parents to engage in the discussion with their child by extending robot's prompts to offer further clarification or concrete examples. Our findings indicate that the parent's desire for the robot to utilize more simple and straightforward language could be based on their desire for the robot to be able to independently be used by their children. However, by providing higher-level and more advanced terms the robot can successfully engage and prompt the parents into delivering math talk to their child.

Overall, the robot impacted parents' level of involvement in the activities by using more advanced math talk concepts, helped parents realize how to incorporate math talk with their child in other activities, and gave them new ideas for what types of math talk to engage with their child.

\subsection{Theme 2: Parents perceived different roles for the robot and preferred for it to independently play with the child}

\paragraph{Parental Involvement} Most parents (R1, R2, R3, R4, R5, R6, R8, R9) preferred for the robot to be able to play with their child independently, while some of them (R1, R3, R5) expressed that they would also want to participate in the interaction occasionally. R2 said that they would prefer for their child to play independently with the robot so that they could ``get some free time.'' R4 echoed a similar sentiment, saying that they could tell their child ``okay I need to make dinner, and you sit and play with your robot.'' R6 added that it would be wonderful to ``find an independent robot that can teach my child math,'' and that it would ``offload some of the responsibility for asking mathematical questions.'' R1 said that the robot ``sort of took the playing off of me a little bit'' and that they felt their child could play with the robot ``independently.''  However, R1 felt that the robot occasionally brought up advanced concepts for their child. While they felt this served as ``a good teaching point,'' they felt the child may ``not be able to do on his own.'' Therefore, some parents (R1, R3, R5) expressed that they would want to be able to ``play all together sometimes,'' and that it ``would be nice to have a mix'' of both independent and shared play capabilities. 

\paragraph{Role of the Robot} Parents thought of the robot as a ``teacher'' (R2, R6, R8), as a ``playmate'' (R1, R4, R5, R9), and as a ``facilitator'' (R1, R3). R6 regarded the robot as a math teacher, saying that they tried to ``leverage Misty's presence as an authority figure'' to motivate the child to address the questions, relieving themself of the ``responsibility for helping to read the book.'' Similarly, R2 felt the robot ``was directing things rather than just giving comments.'' On the other hand, R3 felt that the robot was able to prompt them but that it was less of an authority figure, rather complementing their own abilities and making them realize ``oh yeah I can ask [my child] this.'' R1 noted their perception of the robot's role varied between the two activities, describing the robot as a facilitator in the reading activity and as a playmate in the free-play activity. 

\paragraph{Robot Prompts} Several parents noted a difference in the interaction when the robot gave a question versus a comment (R4, R6, R8), and gave suggestions for how they thought the robot's prompts could be designed (R1, R3, R5, R6, R9). R4 pointed out that when the robot gave a statement, "it's not involving him [the parent]" a lot, whereas a question brought in "an actual interaction where he's being a part of the problem solving." R6 added that they thought the ``statements didn't require an interaction from the child,'' and as a parent, the statements made him feel like he was being tested by the robot as if they ``were supposed to ensure that those statements are accurate to reflect understanding of the material.'' Regarding the robot's prompts, R1 recommended their difficulty could be adjusted as the children age, saying that the robot `` would kind of like grow up with your kid,'' and that it could include other topics such as literacy and colors in order to keep the children more engaged. Moreover, R3 expressed that it was helpful for the robot to provide encouraging feedback, saying she ``could tell that [the child] was happy when [the robot] said `Oh you are awesome.'\thinspace'' 

Overall, parents viewed the robot's role as one that would allow it to primarily engage with the child independently but also wanted it to also be capable of involving the parent occasionally. They also expressed preferences for how the robot could better involve them in a conversation, and the topics include form and variety of prompts, variation of difficulty level, and encouraging feedback.

\subsection{Theme 3: Parents want to have control over the interaction}

\paragraph{Method of Interaction Initiation} Parents expressed three different preferences regarding how the robot \textbf{initiates} the interaction: (1) the parent or child initiates the interaction with the robot (R1, R6), (2) the robot initiates the interaction without any permission (R2, R3, R4, R6), and (3) the robot indicates its intention to initiate the interaction but waits for permission (R1, R2, R3, R5). Some parents (R1, R6) indicated that they prefer the robot to ``let the child actively press a button to get questions,'' as they wanted more control over when to engage with the robot. Other parents (R2, R3, R4, R6) thought that they would prefer the robot to elicit feedback without any prompts or permission, desiring a more natural human-to-human type of interaction. R3 expressed that they hoped the robot ``could speak without prompting,'' and the back-and-forth interaction could ``facilitate their[child and robot] relationship.'' Similarly, R4 described that they ``prefer the robot to be able to release comments without permission, like human conversations,'' which can be ``more beneficial than sitting and waiting for her [the robot] to raise her hand.'' However, R6 suggested they preferred some sort of control over when the robot speaks as they would not like it if the robot would ``just blurt out a question out of the blue.'' 

This desire for control of the interaction was discussed more explicitly regarding the free-play activity, where the robot would raise its hand and wait for permission to talk .Some parents felt that it ``allows them to have the choice to engage or not'' (R3, R5) and that its hand-raising behavior kept the child engaged (R1, R2, R3, R5). Some parents (R1, R2, R5) mentioned that their child pressed the bumper ``as soon as the robot raised her hand,'' which makes the child feel ``excited,'' or that it was ``fun,'' and was perceived to be ``interactive.'' 
However, R3 pointed out that it was ``harder to incorporate her [the robot] when she was not talking,'' and they knew that the robot ``wasn't going to be able to respond.''

\paragraph{Frequency and Timing of Prompts} Several parents (R1, R2, R4, R6) noted how the \textbf{frequency} and \textbf{timing} of the robot's prompts affected their experiences in the interactions. R1 thought the robot's prompts were ``well timed in reading,'' and that having the robot give prompts ``every couple pages was enough'' as ``asking too many questions'' may blur the story. However, R6 noted that showing tags to the robot felt like they were ``taking weird pauses'' in order to receive the robot's questions or statements. R2 felt the process of scanning tags ``was too slow,'' which they felt could make the child ``distracted,'' and R4 noted that slow responses could make ``the parent feel uncertain'' about when the robot would engage in the activity. R4 further explained that they felt like they were ``waiting for her [the robot] to do it [raise its hand],'' but they ``don't know when that moment will come.'' R2 also noted that the reading activity felt more structured, as interaction was determined by showing the tags in the book, while in playing he said it ``depended on what grabbed his[child] focus.''

Overall, parents wanted to have more control over the interaction, but had different preferences for how the robot should engage in the activities and how active of a participant it should be.

\section{Discussion}
Our study explored how the introduction and interaction design of a robot might shape parent-child-robot interaction surrounding math talk. We investigated parents' perceptions of robot-guided math talk through reading and play activities. We discovered that the interactions with the robot \textit{broadened} parental perceptions of math talk with their children and that parents primarily preferred \textit{dyadic} child-robot interactions but wanted more explicit control over the robot and interaction when they were involved. Our results have a number of implications, including that in-home educational robots have the potential to assist and model strategies in parent-child conversations, that parents have preferences for how these robots could share their parenting responsibilities for early education, and that the robot's interaction should prioritize desired parental control. We detail each implication below and conclude with our design recommendations, as summarized in Table \ref{tab:design_recommendations}. Lastly, we discuss the limitations of and several future directions for this work.

\begin{table*}[t!]
    \caption{A summary of the design recommendations for in-home educational robots based on the three themes from the thematic analysis.}
    \label{tab:design_recommendations}
    \centering\renewcommand{\arraystretch}{1.1}
    \begin{tabular}{p{.15\linewidth}p{.375\linewidth}p{.375\linewidth}}
        \toprule
        \textbf{Theme} & \textbf{Recommendation} & \textbf{Example} \\
        \toprule
        \multirow{6}*{Theme 1} & In-home educational robots should be designed as a resource to promote the cognitive thinking of parents and children through effective methods. & The robot can include more advanced terms within its prompts for parents to use and expand on.\\
        \cline{2-3}
         & In-home educational robots should be designed to model or demonstrate learning strategies through social interaction. & The robot could demonstrate to the parent how to scaffold a complex idea or
provide encouragement within the context of the ongoing activity.\\
        \midrule
        \multirow{6}*{Theme 2} & In-home educational robots should be designed to accommodate different levels of parental involvement. & The robot can support the child independently when the parent is busy but also be able to engage or empower parents when they are involved.\\
        \cline{2-3}
         & In-home educational robots need to act and speak in a way that is consistent with the role of the robot. & If the robot's role is to provide instruction like a teacher, it should offer more structure and guidance in the activity.\\
        \midrule
        \multirow{6}*{Theme 3} & In-home educational robots must be designed with the ability to delegate its autonomy to families in a way that meets their expectations. & If there is an ongoing conversation, the robot could communicate its desire to communicate using non-verbal cues and wait until the family members explicitly grant it a turn to speak, giving them control over when the robot engages with the family.\\
        \bottomrule
    \end{tabular}
\end{table*}

\subsection{Robot's cognitive and social methods in promoting parent-child math talk experience}
Parents agreed that the social robot's prompts of math talk facilitated their discussions with their child, which they would not have thought about without the robot. We conclude that the robot was able to assist parent-child math talk by introducing cognitive components to their thinking and recommending topics for the conversations. 
Within this context, the robot served as a cognitive resource providing an implicit educational framework through the use of verbal prompts for parents to discuss with their child, offloading parents' cognitive load in providing the child with math-oriented interaction. 

In addition, parents felt the robot helped them realize new educational strategies for incorporating math talk into other interactions by observing how the child reacted positively to the prompts that the robot offered. This observation implies that the robot potentially influences parent-child math talk through \textit{social learning}. Social learning, a theoretical concept derived from the psychology literature, suggests that humans can learn through imitation when a choice or decision that is carried out by others is perceived as a cost-effective way to attain certain outcomes \cite{bandura1962social}. Through this theoretical lens, the robot could potentially shape or empower parents to carry out more educationally effective interactions with their children. Thus, social learning could be a critical mechanism for social robots to promote parent-child interaction, especially given their physical presence and humanoid designs. Therefore, we provide the following design recommendations:

 \begin{enumerate}
     \item \textbf{In-home educational robots should be designed as a resource to promote the cognitive thinking of parents and children through effective methods.} 
     For example, the robot can include more advanced terms within its prompts for parents to use and expand on.
     \item \textbf{In-home educational robots should be designed to model or demonstrate learning strategies through social interaction.} For example, the robot could demonstrate to the parent how to scaffold a complex idea or provide encouragement within the context of the ongoing activity.
\end{enumerate}


\subsection{Robot's mixed interaction styles and perceived social roles in sharing parent responsibility}

Most parents wished the robot could have been designed in a way that would allow children to be able to play with it independently, as they wanted to save time for other work and thought the robot could reduce some responsibility. This is also reflected by parents' desire for the robot to use simpler and more straightforward language. However, what parents prefer may not have the best educational outcome for their children. Given what we observed in the study as an example, if the robot used more basic concepts and language, parents may not have felt the need to be as involved in the activity. When the robot included advanced or abstract terms and concepts, this created an opportunity for parental math talk as parents were motivated to explain and provide examples for their children. Based on this exploratory observation, the robot's choice of words could influence parents' motivation to participate in the activity and further affect the form of interaction among parent, child, and robot in educational activities. While it is unknown whether the robot's math talk can replace parental math talk in a more effective way, prior work has consistently demonstrated that early parental math talk can predict children's mathematical achievement in preschool \cite{susperreguy2016maternal, levine2010counts, gunderson2011some, eason2020parent, eason2021facilitating}. Since our long-term goal is to design the robot to act as a technology intervention to promote parental math talk at home, it is important to consider whether and how experience with the robot might lead to longer-term durable changes in parent behavior, even after the robot has been removed. For those changes to occur, parents must observe the robot modeling math talk with their children. 


Moreover, the perceived role of the robot differs across families and could influence parents' social behavior in parent-child-robot interaction by changing how they incorporate and use the robot. Based on our interviews, the robot's perceived role was described differently, as a facilitator, as a playmate, or as a teacher. A playmate and a teacher are two significantly different roles, showing that parents have different perceptions of the robot's role even though the robot provided the exact same interaction for all families. One possible explanation is that parents have different \textit{initial expectations} regarding the robot's responsibility, biasing their viewpoints of the robot's intention to give the prompts. 
For example, parents who viewed the robot as a teacher tried to use its perceived authority as a means of getting children to answer questions with discipline and let the robot take on more responsibility to motivate the child. Whereas, those who perceived it as a playmate were able to encourage the child in a joyful attitude and be more accountable for interacting with ``two kids.'' These differing perceptions that parents have of the robot could have a great impact on how they introduce the robot to their child, how the child gets to know who the robot is, and what attitude they decide to take with the robot's prompts. 
Thus, when designing a social robot for parent-child dyads, it is important to consider the alignment between the design for the robot's role and the users' perception. We recommend the following specific recommendations:

 \begin{enumerate}
     \item \textbf{In-home educational robots should be designed to accommodate different levels of parental involvement.} For example, robots can support the child independently when the parent is busy but also be able to engage or empower parents when they are involved. 
     \item \textbf{In-home educational robots need to act and speak in a way that is consistent with the role of the robot.} For example, if the robot's role is to provide instruction like a teacher, it should offer more structure and guidance in the activity.
\end{enumerate}


\subsection{Robot's multi-level autonomy in supporting personalized preferences for interaction}
From our interviews, parents discussed three methods for the robot to prompt interaction---\textit{user-controlled}, \textit{autonomous}, and \textit{indication}---each with a different amount of control for the parent or child to have in initiating the interaction. 
The first method, \textit{user-controlled}, allows for the most control over the interaction and involves the child or the parent actively prompting the robot in order to interact with it. However, their experience interacting with the robot may not be natural, and families may be more likely to disengage from a robot using such a system. The second method, \textit{autonomous}, allows the robot to act fully autonomously and allows the robot to initiate the interaction without direct permission. Parents thought that the \textit{autonomous} would be more similar to a human-like interaction and would permit them to develop a closer relationship with the robot. However, some parents expressed concerns about being interrupted by the robot. The last method, \textit{indication}, is similar to the \textit{autonomous} one but gives families more control over when to engage the robot. This method still has the robot determining when it wants to initiate the interaction but has it give an indication that it wants to do so, such as raising its hand, and waiting for the participant's permission to engage. For this method of interaction, parents indicated that they liked having the choice of whether to involve the robot and that when they did the robot's prompting behavior kept them engaged. However, as shown in our results, the \textit{indication} method can make it difficult for parents to include the robot in the activity, as it can make them feel uncertain about when the robot would interact next. Therefore, we identified the following design recommendation:

 \begin{enumerate}
     \item \textbf{In-home educational robots must be designed with the ability to delegate its autonomy to families in a way that meets their expectations.} For example, rather than interrupting an ongoing conversation, the robot could communicate its desire to communicate using non-verbal cues and wait until the family members explicitly grant it a turn to speak, giving them control over when the robot engages with the family.
\end{enumerate}

\subsection{Limitations \& Future Directions}
Our work has a number of limitations. First, our study involved a small sample of nine participants. This small sample size is due to the exploratory nature of our study, our choice of an in-home field setting, and the challenges in recruiting families with children in a narrow age range. Second, we recruited participants via a university mailing list, presumably leading to a more highly educated, and motivated sample. Moreover, the study is conducted in the United States and within two specific contexts (i.e., reading and free play), which may limit the findings to certain cultural contexts and activities.

Our study utilized a Wizard-of-Oz (WoZ) approach to engage the robot in the tasks with the families, and further exploration is necessary to determine whether robots can autonomously engage in math talk in a reliable fashion and whether the autonomous robot elicits similar outcomes. Our preliminary findings motivate us to develop an autonomous robot system that can facilitate math talk and study its effects in a long-term in-home deployment study. Such a study will allow us to use standardized mathematics assessments before and after the intervention to evaluate growth in children's math skills, which we did not include in the current study given the brevity of the interactions. This planned study would also address questions regarding long-term use and whether such a system can actually affect both parents' habits in delivering math talk and their ability to transfer the skill to incorporate math talk in different activities without the robot's assistance. 



\section{Conclusion}
In this paper, we explored the use of a social robot in promoting ``math talk'' within parent-child interactions to gain an understanding of the role of a social robot in a triadic social dynamic. We designed an autonomous \textit{reading} and Wizard-of-Oz (WoZ) \textit{free-play} activities in which a robot used a series of carefully timed prompts to promote parental use of math talk. 
Our qualitative findings offer insights into how families were inspired by the robot's prompts; their desired interaction style and method for the robot; and how they wanted to include and engage with the robot during the interaction. Informed by these results, we provide several design recommendations for in-home math-focused social robots, as well as several avenues for future research.

\section{Selection and Participation of Children}
This study stuck to an ethical protocol that was reviewed and approved by the \texttt{University of Wisconsin --- Madison} Institutional Review Board (IRB) under protocol \texttt{2020-1264}. After the protocol was reviewed and accepted, children were recruited via their parents who were contacted through organizational mailing lists. During the consent process, the experimenter first explained the study procedure and then had the parent fill out a written consent form. The experimenter then informed the child about the study and obtained verbal assent. After acquiring the child's verbal assent, the experimenter began the study. Participants were informed that collected data would only be shared with members of the study team and that any pictures or videos would be anonymized prior to publication. Parents were also notified that the experimenter would break confidentiality only if abuse or neglect was determined during the study. Participants received \$15 USD compensation for participating in the study.

\begin{acks}
We would also like to thank our undergraduate research assistants, Andreea Turcu, Jack Siepmann, and Abby Andrews for supporting the qualitative coding process.
\end{acks}

\bibliographystyle{ACM-Reference-Format}
\bibliography{sample-base}


\end{document}